\documentclass{ecai}
\usepackage{graphicx}
\usepackage{latexsym}
\usepackage{times}
\usepackage{latexsym}
\usepackage{amsmath}
\usepackage{amsfonts}
\usepackage{multirow}
\usepackage{textcomp}
\usepackage{booktabs}
\usepackage{bm}
\usepackage{color}
\usepackage{setspace}
\usepackage{array}
\usepackage{stfloats}
\usepackage{url}
\usepackage{graphicx}
\usepackage{mathbbol}
\usepackage{enumitem}
\usepackage{cancel}
\usepackage[T1]{fontenc}
\newcommand{\ignore}[1]{}

\usepackage[utf8]{inputenc}

\usepackage{microtype}

\usepackage{inconsolata}

\begin{document}

\begin{frontmatter}

\title{Constrained Multi-Layer Contrastive Learning \\for Implicit Discourse Relationship Recognition}

\author[A]{\fnms{Wu}~\snm{Yiheng}}
\author[B]{\fnms{Li}~\snm{Junhui}}
\author[C]{\fnms{Zhu}~\snm{Muhua}}

\address[A]{Aian.online}
\address[B]{Soochow University}
\address[C]{Meituan}
\begin{abstract}
Previous approaches to the task of implicit discourse relation recognition (IDRR) generally view it as a classification task. Even with pre-trained language models, like BERT and RoBERTa, IDRR still relies on complicated neural networks with multiple intermediate layers to proper capture the interaction between two discourse units. As a result, the outputs of these intermediate layers may have different capability in discriminating instances of different classes. To this end, we propose to adapt a supervised contrastive learning (CL) method, label- and instance-centered CL, to enhance representation learning. Moreover, we propose a novel constrained multi-layer CL approach to properly impose a constraint that the contrastive loss of higher layers \ignore{in the baseline model }should be smaller than that of lower layers. Experimental results on PDTB 2.0 and PDTB 3.0 show that our approach can significantly improve the performance on both multi-class classification and binary classification. 
\end{abstract}

\end{frontmatter}

\section{Introduction}
Discourse relation recognition (DRR) aims to identify the existence of logical relations between two adjacent text spans in a discourse. Such text spans are called discourse units (DUs) which can be clauses, sentences, and paragraphs. Advance in this task is beneficial to many downstream tasks, such as text summarization ~\cite{Geran_etal_2014_Abstractive} and question answering ~\cite{Jansen_etal_2014_Discourse}. Depending on the existence of connectives in DUs, DRR generally falls into two broad categories: explicit discourse relation recognition (EDRR) and implicit discourse relation recognition (IDRR). Since connectives provide informative linguistic cues for recognizing discourse relations, straightforward application of general classifiers can achieve above $90\%$ accuracy for EDRR ~\cite{Pitler_etal_2009_Using, Pitler_etal_2008_Easily}. By contrast, due to the absence of connectives, IDRR can utilize the semantics of two DUs only, which makes it a challenging task. In this paper we put focus on the task of IDRR. 

Conventional approaches to IDRR put emphasis on the engineering of hand-crafted linguistic features from DUs~\cite{Pitler_etal_2009_Using, Lin_etal_2009_Recognizing}. Even with development of pre-trained language models such as BERT~\cite{Devlin_etal_2019_BERT} and RoBERTa~\cite{Liu_etal_2019_RoBERTa}, recent advances in IDRR still rely on complicated neural networks with multiple intermediate layers to proper capture the interaction between two discourse units. \ignore{More recently, with the development of contextualized word representation learning such as BERT~\cite{Devlin_etal_2019_BERT} and RoBERTa~\cite{Liu_etal_2019_RoBERTa}, an amount of effort has been put on the design of complicated neural network structures that integrate large pre-trained models and sophisticated task-specific modules. }For example, Ruan et al.~\cite{Ruan_etal_2020_Interactively} emphasize the importance of interactive-attention for IDRR. Liu et al.~\cite{Liu_etal_2020_On} claim that a bilateral multi-perspective matching module and a global information fusion module are also important to this task. \ignore{\textcolor{red}{These well-designed sophisticated models are effective on language understanding at both the text span level and the sentence level. However, these models also introduce more parameters and lead to high complexity.}} Taking the model of Liu et al.~\cite{Liu_etal_2020_On} as an example, however, our close examination reveals that as expected, the outputs of its intermediate layers have different capability in discriminating instances of different classes. To this end, \ignore{In this work, we aim to build a relatively simple yet effective model for IDRR. For this purpose, we first build a baseline model which is a simplified version of ~\cite{Liu_etal_2020_On}. Meanwhile, to compensate for the loss of learning capability incurred by model simplification, }we propose to adapt supervised contrastive learning (CL) to enhance representation learning of these intermediate layers. It is worthy noting that the use of supervised CL does not introduce additional parameters during the inference phase. 
 
In recent years, supervised CL has been widely applied in various natural language processing tasks~\cite{Suresh_etal_2021_Not, Gunel_etal_2021_Supervised}. The basic idea of supervised CL is to pull semantically close neighbors together, and to push non-neighbors away~\cite{Gao_etal_2021_SimCSE}. Different from them, in this paper we propose a novel constrained multi-layer contrastive learning (CMCL) approach for IDRR. Specifically, we first design a simple yet effective CL method, label- and instance-centered contrastive learning (LICL), which is an extension of label-centered contrastive learning (LCL)~\cite{Zhang_etal_2022_Label}. We then plug LICL into each layer of our baseline model, i.e., multi-layer CL. Finally, to alleviate the redundancy introduced by applying LICL in multiple consecutive layers, we propose CMCL which further imposes a constrain that the LICL loss of higher layers should be smaller than that of lower layers. Experimental results on PDTB 2.0 and PDTB 3.0 show that applying LICL to any intermediate layer achieves significant improvement over the baseline model. Moreover, applying LICL to multiple intermediate layers with the proposed constrained multi-layer CL approach further improve the performance.  

\section{Motivation}
Thanks to large-scale pre-trained models, the performance of IDRR has made a breakthrough~\cite{Wu_etal_2020_Hierarchical, Liu_etal_2020_On}. 
However, different from sentence-level NLP tasks including sentiment analysis~\cite{Mercha-etal-2023-Machine} and fallacy recognition~\cite{Alhindi-etal-2022-multitask}, for which it only requires lightweight networks upon pre-trained models, IDRR models in recent studies rely on complicated networks which generally consist of multiple intermediate layers. For example, from the bottom to the top the model in~\cite{Liu_etal_2020_On} is composed of a contextualized representation layer, a bilateral multi-perspective matching layer, and a global information fusion layer which are critical to achieve good performance. In this paper we aim to better apply CL to a typical IDDR model which consists of multiple layers. In the following, we first describe the network structure of our baseline model which is a simplified version of \cite{Liu_etal_2020_On}, and then provide analysis to demonstrate the potential of using CL to enhance the representation learning. 
\begin{figure}[!t]
\centering
\includegraphics[width=3.2in]{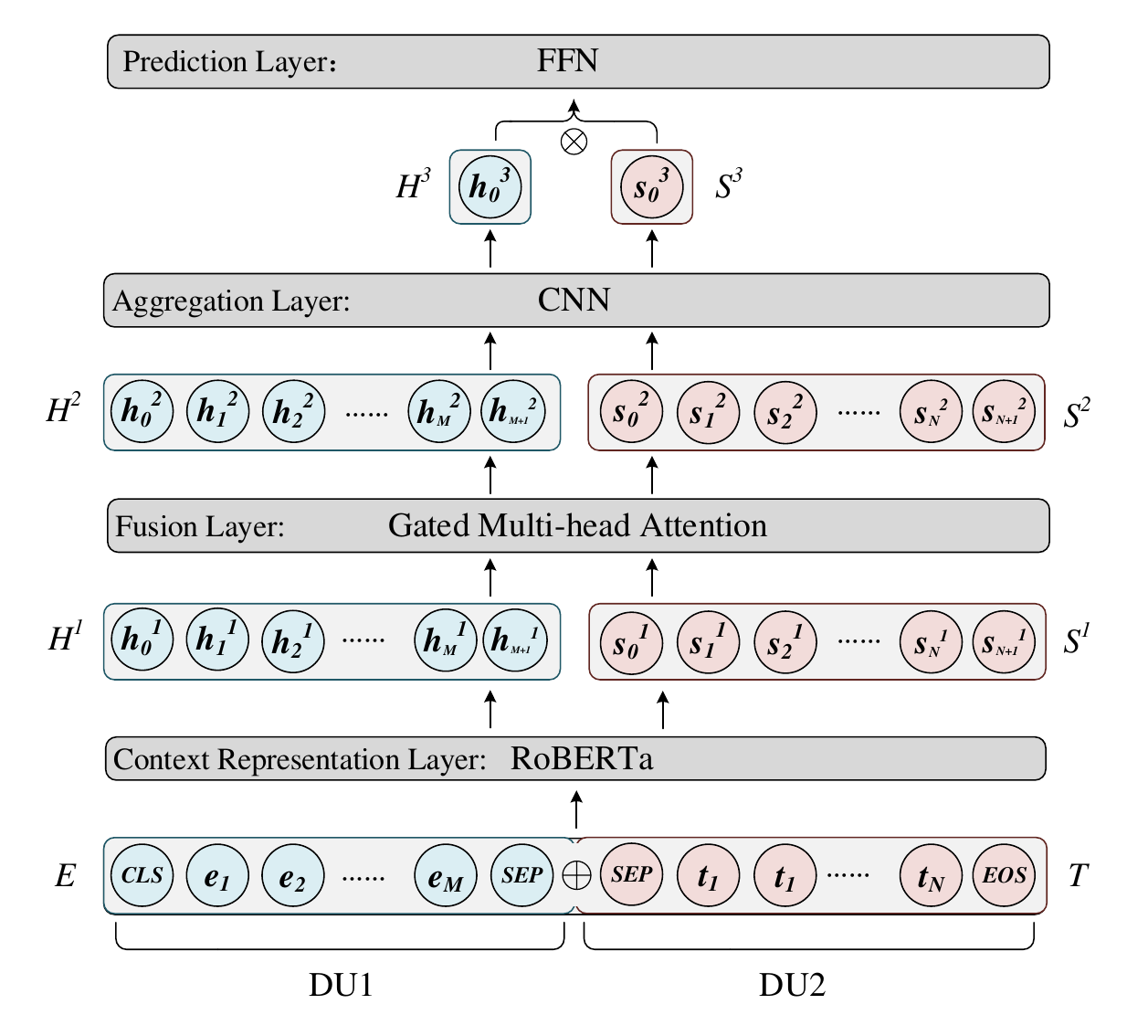}
\caption{Architecture of our baseline model.}
\label{fig:model_overview}
\end{figure}

\subsection{Baseline Model}
Figure~\ref{fig:model_overview} presents our baseline model which is a simplified version of \cite{Liu_etal_2020_On}. \ignore{\textcolor{red}{We only take advantage of its backbone to build an IDRR model as simple as possible and eliminate potential influencing factors. }}The baseline model is composed of the following layers from the bottom to the top. We direct the reader to \cite{Liu_etal_2020_On} for more details.

\noindent\textbf{Context Representation Layer.}
Texts in DUs are encoded into embedding representations by using RoBERTa. In order to fully utilize contextualized understanding of RoBERTa, we concatenate a pair of DUs with a special token {\texttt{[SEP]}} in between and then obtain contextualized representations by splitting it for them:
\begin{equation}
\small
\left(H^1,S^1\right) = \text{RoBERTa}\left([E, \texttt{[SEP]}, T]\right),
\end{equation}
where {\small $E=(e_1,\cdots,e_M)$}, {\small $T=(t_1,\cdots,t_N)$} are input texts of DU1 and Du2, respectively; {\small $M$}, {\small $N$} are length of {\small $E$}, {\small $T$}.\footnote{The input of RoBERTa is the addition of token embeddings, position embeddings, and segment embeddings. The segment embeddings view DU1 and DU2 differently and are learnable~\cite{Liu_etal_2020_On}.}

\noindent\textbf{Fusion Layer.} Fusion layer is used to further abstract semantic information among individual DUs. It is a vanilla Transformer layer by replacing layer normalization with a gated mechanism. Taking DU1 as example, it takes {\small $H^1\in\mathbb{R}^{(M+2)\times d_1}$} as input and computes {\small $H^2\in\mathbb{R}^{(M+2)\times d_2}$} via a gated multi-head attention function~\cite{Liu_etal_2020_On}:
\begin{equation}
\small
H^2 = \text{GatedMHA}\left(H^1W^Q, H^1W^K, H^1W^V\right),
\end{equation}
where {\small $W^Q, W^K, W^V \in \mathbb{R}^{d_1\times d_2}$} are model parameters to reduce the dimension from {\small $d_1$} to {\small $d_2$}. Likewise, we get {\small $S^2\in\mathbb{R}^{(N+2)\times d_2}$} for DU2. 

\noindent\textbf{Aggregation Layer.} 
We leverage a series of convolutional neural networks (CNNs) followed by a highway network to capture 1-gram, {\small$\cdots$}, {\small $J$}-gram information. Then we apply maximum pooling to obtain representations with fixed dimensions for the prediction layer.
\begin{equation}
\small 
o_j = \text{ReLU}\left(\text{max}\left(\text{Conv}_j(H^2)\right)\right),
\end{equation}
\begin{equation}
\small
o = \text{Flatten}\left([o_1,\cdots,o_J]\right),
\label{equ:conv}
\end{equation}
\begin{equation}
\small
H^3 = \text{Highway}\left(o\right),
\end{equation}
where {\small $o_j \in \mathbb{R}^{d_3}$} is the output of the {\small $j$}-th convolution {\small $\text{Conv}_j$}. We utilize kernels whose sizes vary from 1 to {\small $J$} to extract 1-gram, {\small$\cdots$}, {\small $J$}-gram information, and all output dimensions of convolutions are set to {\small $d_3$}. {\small $\text{Flatten}([\cdots])$} indicates flatten concatenation of all outputs from convolutions while {\small $\text{Highway}(\cdot)$} indicates residual connection combined with two gates~\cite{Srivastava_etal_2015_Highway}. {\small $H^3\in \mathbb{R}^{Jd_j}$} is the output vector of aggregation layer. Similarly, we get {\small $S^3\in \mathbb{R}^{Jd_j}$} for DU2.

\noindent \textbf{Prediction Layer.}
Given {\small $H^3$} and {\small $S^3$}, the prediction layer is a single-layer feed-forward neural network to obtain the probability distribution. For the label set {\small $C$}, its label can be estimated as:
\begin{equation}
\small
p(\Tilde{y}) = \text{softmax}\left([H^3, S^3](W^\theta)^{T}\right),
\end{equation}
where {\small $W^\theta \in \mathbb{R}^{|C|\times(2Jd_3)}$} is a trainable matrix. We optimize the baseline model with cross entropy loss between {\small $\Tilde{y}$} and the ground-truth label $y$.

\subsection{Analysis}
As shown in Figure~\ref{fig:model_overview}, the baseline model consists of multiple consecutive layers between the input layer and the output layer. We conjecture that the outputs of these intermediate layers have different capability in discriminating instances of different classes. To this end, we visualize the representations of instances using the outputs of the context representation layer, the fusion layer, and the aggregation layer, respectively. Specifically, given an instance, we concatenate the outputs of its DU1 and DU2 as the representation of the distance.\footnote{The representation of (DU1, DU2) pair is {\small $[s_0, h_0]$} in Eq.~\ref{equ:mu}.} We then follow previous works~\cite{Gunel_etal_2021_Supervised, Zhang_etal_2022_Label} to visualize instance representations with T-SNE on a 2D map \cite{Laurens_etal_2008_visualizing}.

\begin{figure*}
\centering
\includegraphics[width=6in]{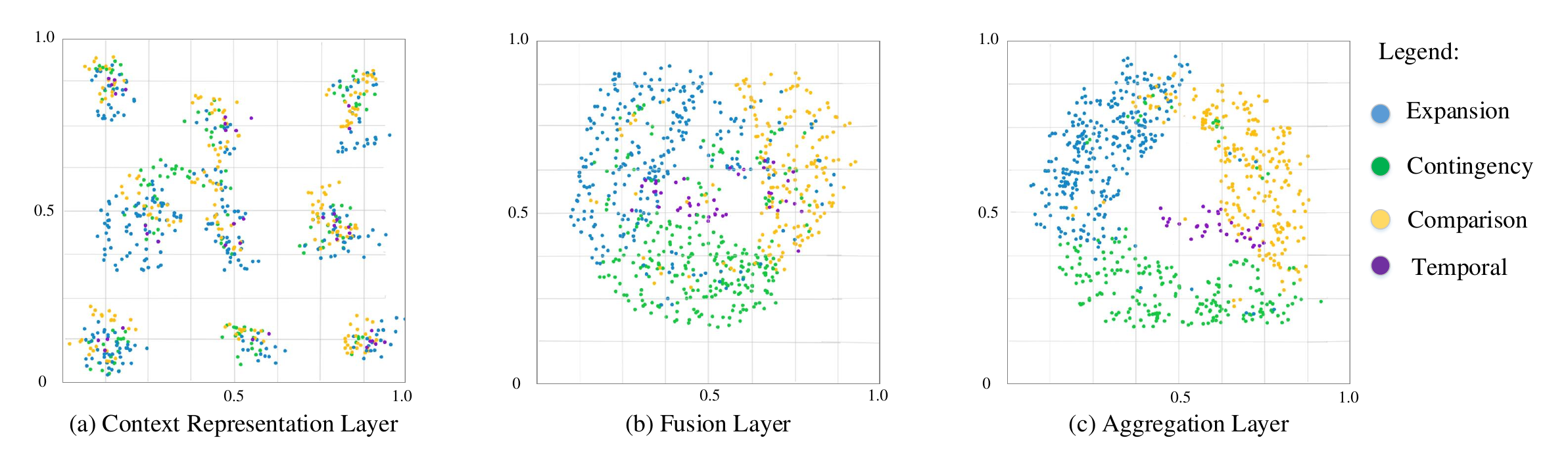}
\caption{Visualization on representations of the baseline model on PDTB 2.0 development set using T-SNE.}
\label{fig:baseline_tsne}
\end{figure*}

Figure~\ref{fig:baseline_tsne} shows the visualization on representations of the baseline model on PDTB 2.0 development set. As expected, the representations from the highest layer (i.e., the aggregation layer, Figure~\ref{fig:baseline_tsne} (c)) have the best capability in discriminating different instances while the representations from the lowest layer (i.e., the context representation layer, Figure~\ref{fig:baseline_tsne} (a)) are confusing and disorganized. It also shows that the instances of \textit{Temporal} are most difficult to recognize from the instances of other classes, probably due to its very limited training instances. 

Considering the big capability gap between the representations from the lowest and the highest layers, this motivates us to narrow the gap by explicitly enhancing the capability of representation learning of lower layers in discriminating instances of different classes. We believe that higher layers could also benefit from the enhanced capability of lower layers, which in turn will improve the final classification performance.

\section{Our Approach}
\label{sect:approach}
In this section, we first introduce label-centered contrastive learning (LCL) (Section~\ref{sect:lcl}), which we couple with instance-centered contrastive learning and propose label- and instance-centered contrastive learning (LICL) (Section~\ref{sect:licl}). Then we apply LICL to the representations of multiple layers and propose constrained multi-layer contrastive learning (CMCL) (Section~\ref{sect:CMCL}). 

\subsection{LCL: Label-centered Contrastive Learning}
\label{sect:lcl}
We adapt the label-centered contrastive learning approach (LCL) proposed by Zhang et al.~\cite{Zhang_etal_2022_Label} in our model. Given a batch {\small $\mathcal{S}$} with {\small $N$} input instances and corresponding labels {\small $\mathcal{S}=\{\left(x_i, y_i\right)\}|_{i=1}^N$}, we define instances in this batch whose label is not {\small $c$} as {\small $B\left(c\right)$}, i.e., {\small $B(c) = \{x_i|y_i\neq c\}$}. For the $i$-th instance and its label {\small $\left(x_i, y_i\right)$}, our optimization objective is to maximize:
\begin{equation}
\small
\text{Pos}_{\texttt{LCL}}\left(i\right)=\exp{\left(Sim\left(L_{y_i}, \mu_{x_i}\right)/\tau\right)},
\label{equ:pos1}
\end{equation}
where {\small $L_{y_i}\in\mathbb{R}^{d}$} is label vector for label {\small $y_i$} and {\small $\mu_{x_i}\in\mathbb{R}^{d}$} is a representation for instance {\small $x_i$}, and $\tau$ is the temperature hyper-parameter, which can be tuned on a development set; {\small $Sim\left(\cdot, \cdot\right)$} is a cosine similarity function used for measuring semantic similarity. It also tries to minimize:
\begin{equation}
\small
\text{Neg}_{\texttt{LCL}}\left(i\right)=\sum_{x\in B\left(y_i\right)}\exp{\left(Sim\left(L_{y_i}, \mu_{x}\right)/\tau\right)}.
\label{equ:neg1}
\end{equation}

Generally, the ultimate optimization objective of LCL is composed of these two functions and defined as:
\begin{equation}
\small
\text{Loss}_{\texttt{LCL}}\left(\mathcal{S}\right)=-\frac{1}{|C|}\sum_{i=1}^{N}\gamma_{y_i}\log\frac{\text{Pos}_{\texttt{LCL}}\left(i\right)}{\text{Neg}_{\texttt{LCL}}\left(i\right)},
\end{equation}
\begin{equation}
\small
\gamma_{y_i} = \frac{avg\left(\sum_{c_i\in C}|D_{c_i}|\right)}{|D_{y_i}|},
\end{equation}
where {\small $C$} is the label set, {\small $|D_{c}|$} indicates the total number of training instances with class $c$; {\small $\gamma_{y_i}$} is a trade-off weight to overcome the imbalance of training instances.

In above, each instance {\small $x$} consists of two DUs. We first obtain its representation {\small $\mu_{x}$} via: 
\begin{equation}
\small
\mu_{x} = ([h_0,s_0])W^{\mu},
\label{equ:mu}
\end{equation}
where {\small $h_0\in\mathbb{R}^{d}$} and {\small $s_0\in\mathbb{R}^{d}$} are representations of DU1 and DU2, respectively, {\small $W^{\mu} \in \mathbb{R}^{2d \times d}$} is a trainable parameter matrix, $[\cdot, \cdot]$ denotes concatenation, and {\small $d$} is the size of representation whose value is dependent on the layer used. Specifically, as shown in Figure~\ref{fig:model_overview}, we follow Liu et.al ~\cite{Liu_etal_2020_On} and take the first special token (i.e., {\texttt{[CLS]}} and {\texttt{[SEP]}}) as representative of DU1 and DU2, respectively. Besides, the representation of DU1 and DU2 can be extracted from the output of either the context representation layer, the fusion layer, or the aggregation layer.

 As shown in Figure~\ref{fig:CMCL} (a), LCL takes label as center, and encourages the label vector to be more similar to the corresponding instances belonging to the same class (i.e., the solid lines in the figure) in a batch than the instances with different labels (i.e., the dashed lines in the figure).  

\subsection{LICL: Coupling Label- and  Instance-centered Contrastive Learning}
\label{sect:licl}

LCL defined in Section~\ref{sect:lcl} explicitly tries to discriminate positive instances from negative instances. However, it does not learn to discriminate positive label from negative labels. Therefore, we propose to couple label- and instance-centered contrastive learning (LICL), in which the later, i.e., instance-centered contrastive learning (ICL) aims to encourage each text representation and corresponding label vector to be closer while pushing far away mismatched instance-label pairs, as shown in Figure~\ref{fig:CMCL} (b). For the $i$-th instance and its label {\small $\left(x_i, y_i\right)$}, the ICL optimization objective is to maximize: 
\begin{equation}
\small
\text{Pos}_{\texttt{ICL}}\left(i\right)=\exp{\left(Sim\left(L_{y_i}, \mu_{x_i}\right)/\tau\right)},
\label{equ:Pos2}
\end{equation}
which is the same as LCL, as in Eq.~\ref{equ:pos1}. ICL also tries to minimize:
\begin{equation}
\small
\text{Neg}_{\texttt{ICL}}\left(i\right) = \exp{\left(Sim\left(L_{c*}, \mu_{x_i}\right)/\tau\right)},
\label{equ:neg2}
\end{equation}
where {\small $c^{*}$} is the negative label with the biggest similarity with {\small $x_i$},\footnote{See Section \ref{sect:experiment_settings} for discussion if we define {\small $\text{Neg}_{\texttt{ICL}}$} as {\small $\sum_{c\in C\&\&c\neq y_i}\exp\left(Sim\left(L_{c}, \mu_{x_i}\right)\right)$}.} i.e.,
\begin{equation}
\small
c^{*} = \operatorname*{arg\,max}_{c\in C \&\& c\neq y_i} Sim\left(L_{c}, u_{x_i}\right).
\label{equ:p_star}
\end{equation}

Finally, given the batch {\small $\mathcal{S}$}, the ultimate optimization objective of LICL is defined as:
\begin{equation}
\small
\text{Loss}_{\texttt{LICL}}\left(\mathcal{S}\right) = -\frac{1}{|C|}\sum_{i=1}^{N}\gamma_{y_i}\log\frac{1}{2}\times\frac{\text{Pos}_{\texttt{LCL}}\left(i\right) + \text{Pos}_{\texttt{ICL}}\left(i\right)}{\text{Neg}_{\texttt{LCL}}\left(i\right) + \text{Neg}_{\texttt{ICL}}\left(i\right)}.
\label{equ:licl}
\end{equation}

\subsection{CMCL: Constrained Multi-Layer Contrastive Learning}
\label{sect:CMCL}
With LICL, it is a straightforward way to apply it to each intermediate layer of the baseline model in Figure~\ref{fig:model_overview}. However, simply and independently applying LICL to multiple layers brings redundancy which may in turn weaken its effect. Thus we propose constrained multi-layer contrastive learning (CMCL).
  
\begin{figure*}
\centering
\includegraphics[width=6in]{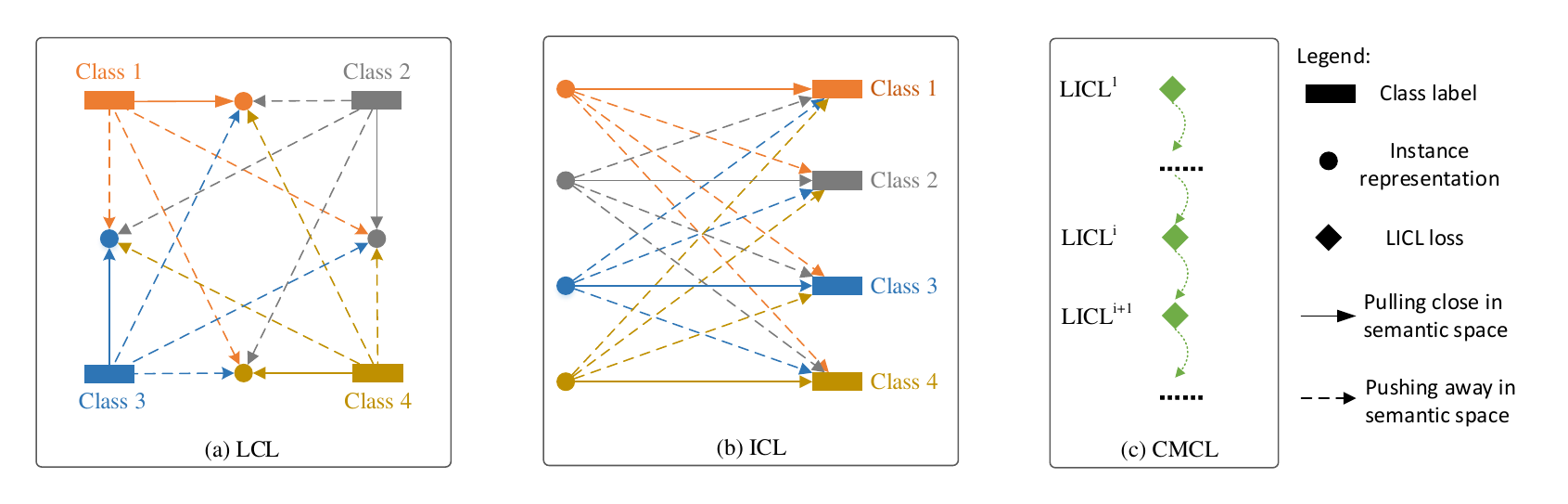}
\caption{Illustration of LCL, ICL, and CMCL.}
\label{fig:CMCL}
\end{figure*}

Without loss of generality, we assume that an IDRR model consists of {\small $L$} layers. As shown in Figure~\ref{fig:CMCL} (c), we use {\small $\text{LICL}^{i}$} to indicate that we apply LICL on the representation learning of the $i$-th layer. To meet the expectation that the representations of higher layers should be more discriminative than those of lower layers, the basic idea of CMCL is that the LICL loss of higher layers should be smaller than that of lower layers. Therefore, given a batch {\small {$\mathcal{S}$}}, the loss of CMCL could be defined as:
\begin{equation}
\small
\begin{split}
\text{Loss}_{\texttt{CMCL}}\left(\mathcal{S}\right)=\sum_{i=1}^{L-1}\max(0, \text{Loss}_{\texttt{LICL}}^{i+1}\left(\mathcal{S}\right) -\text{Loss}_{\texttt{LICL}}^{i}\left(\mathcal{S}\right)-\eta),
\end{split}
\label{equ:cmcl}
\end{equation}
where {\small $\text{Loss}_{\texttt{LICL}}^{i}\left(\mathcal{S}\right)$} is the loss of LICL when it is applied on the representation learning of the $i$-th layer, and $\eta$ is a margin hyper-parameter.

Moving back to our IDRR model as shown in Figure~\ref{fig:model_overview}, we use {\small $\text{LICL}^{1}$}, {\small $\text{LICL}^{2}$}, and {\small $\text{LICL}^{3}$} to indicate that the LICL is applied on the representation learning of the context representation layer, the fusion layer, and the aggregation layer, respectively.  

\subsection{Training Objective}
For batch {\small $\mathcal{S}$}, our CMCL model tries to optimize  the following joint training objective:
\begin{equation}
\label{equ:final_loss}
\small 
\text{Loss}\left(\mathcal{S}\right) = \text{Loss}_{\texttt{CE}}\left(\mathcal{S}\right) + \text{Loss}_{\texttt{CMCL}}\left(\mathcal{S}\right) + \lambda\text{Loss}_{\texttt{LICL}}^{1}\left(\mathcal{S}\right),
\end{equation}
where {\small $\text{Loss}_{\texttt{CE}}\left(\mathcal{S}\right)$} is the cross entropy (CE) loss on {\small $\mathcal{S}$}, and $\lambda$ is a trade-off hyper-parameter to balance the influence of loss of LICL applied at the output of context representation layer.\footnote{In Eq.\ref{equ:final_loss}, we set the weight of {\small $\text{Loss}_{\texttt{CMCL}}\left(\mathcal{S}\right)$} as 1.0 to reduce the computation load in tuning hyper-parameters.}

\begin{table}
\centering
\small
\begin{tabular}{l|l}
\hline
\bf Hyper-Parameter & \bf Value\\
\hline
Learning rate & 0.001\\
Maximum training epoch & 50\\
Batch size & 24\\
Random seed & 999\\
Optimizer & Adam\\
Maximum length of input & 512 \\
\hline
\end{tabular}
\caption{Hyper-parameter values in our experiments.}
\label{tab:parameter_settings}
\end{table} 

\section{Experimentation}
\subsection{Experimental Settings}
\label{sect:settings}
\textbf{Datasets}. We carry out experiments on two benchmarks: Penn Discourse Treebank 2.0 (PDTB 2.0)~\cite{Prasad_etal_2008_penn} and Penn Discourse Treebank 3.0 (PDTB 3.0)~\cite{Webber_etal_2019_penn}. PDTB 2.0 has three levels of senses, i.e., classes, types, and sub-types. Similar as previous studies, we evaluate our approach on both the 4 top-level implicit senses (4-way) and the 11 major second-level implicit types (11-way). The top-level implicit senses are \textit{comparison (Comp.)}, \textit{contingency (Cont.)}, \textit{expansion (Exp.)}, and \textit{temporal (Temp.)}. We use the conventional data splitting, i.e., sections 2-20 for training, sections 0-1 for system development, and sections 21-22 for testing~\cite{Ji_etal_2015_one}. PDTB 3.0 is an upgraded version of PDTB 2.0, which has the same set of top-level implicit senses as PDTB 2.0, but has quite different sub-relations. Following recent studies on PDTB 3.0, we evaluate our approach only on the 4 top-level implicit senses. We use the same data splitting for PDTB 3.0 as that for PDTB 2.0.

\noindent\textbf{Metrics.} Macro-averaged F1 and accuracy (Acc) are used as metrics for both 4-way and 11-way classification. We report F1 for binary classification, which comes from one-versus-all strategy on the 4-way classification. 

\noindent \textbf{Configuration}. We use RoBERTa-base as the context representation layer which is not fine-tuned. Our implementation is based on Huggingface Transformers.\footnote{\url{https://github.com/huggingface/transformers}} We conduct a grid-based hyper-parameter sweep for $\lambda$ between 0.1 (0.01) and 1.0 (0.1) with step 0.1 (0.01) on the development set of PDTB 2.0 (PDTB 3.0). The dimensions of $d_1$, $d_2$, and $d_3$ are set to 768, 128, and 64, respectively. The number of filters in the aggregation layer is set to 64. We set {\small $J$} to 2, and follow Pan et al.~\cite{Pan_etal_2021_Contrastive} and set $\tau$ to 1.0. {\small $\eta$} is set to 0.02. 
Table~\ref{tab:parameter_settings} shows the hyper-parameters of our experiments. When training, CMCL loss {\small $\max(0, \text{Loss}_{\text{LICL}}^{2}\left(\mathcal{S}\right)-\text{Loss}_{\text{LICL}}^{1}\left(\mathcal{S}\right)-\eta)$} is only used to update parameters in the fusion layer while {\small $\max(0, \text{Loss}_{\text{LICL}}^{3}\left(\mathcal{S}\right)-\text{Loss}_{\text{LICL}}^{2}\left(\mathcal{S}\right)-\eta)$} is used to update parameters in the aggregation layer. All our experiments run on a V100 GPU card with 32GB memory. We run our models once with a fixed seed.

\begin{table*}
\small
\centering
\begin{tabular}{l|cc|cc|cc}
\hline
\multirow{3}{*}{\bf Model} & \multicolumn{4}{c|}{\bf PDTB 2.0}&\multicolumn{2}{c}{\bf PDTB 3.0} \\
\cline{2-7}
 & \multicolumn{2}{c|}{\bf 4-way} &\multicolumn{2}{c|}{\bf 11-way}&  \multicolumn{2}{c}{\bf 4-way}\\
& \bf Acc & \bf F1 &\bf  Acc &\bf F1& \bf Acc &\bf F1\\
\hline
Lan et al.~\cite{Lan_etal_2017_Multi}{\small $^\dagger$} & -& -& -&-& 57.06& 47.29 \\
Nguyen et al.~\cite{Nguyen_etal_2019_Employing}{\small $^\ddagger$}& 68.42& 61.89 &57.72 & 38.10 & -& -\\
Wu et al.~\cite{Wu_etal_2020_Hierarchical}{\small $^\ddagger$} &70.05 &62.02&	58.61 &38.28 &-&-\\
Ruan et al.~\cite{Ruan_etal_2020_Interactively}{\small $^\dagger$} & -& -& -&-& 58.01 & 49.45\\
Liu et al.~\cite{Liu_etal_2020_On}{\small $^\bullet$} & 69.06 & 63.39 & 58.23 & 40.02 &	69.88 & 65.15\\
Dou et al.~\cite{Dou_etal_2021_CVAE} &70.17&65.06&-&-&-&-\\
Wu et al.~\cite{Wu_etal_2022_A}  &\underline{71.18} &63.73&60.33& 40.49 &-&-\\
Xiang et al.~\cite{Xiang_etal_2022_Encoding}& -&- &- &- &64.04& 56.63\\
Zhou et al.~\cite{Zhou-etal-2022-prompt-based}&70.84 &64.95&60.54 & 41.55 &-&- \\ 
Long et al. \cite{Long-etal-2022-facilitating}& \textbf{72.18}& \textbf{69.60} & \textbf{61.69} &\textbf{49.66}&\textbf{75.31} &\textbf{70.05}\\
\hline
Baseline (B) &67.59&62.19&56.98& 39.59 &68.24& 64.26\\
B + CMCL& 70.65&\underline{65.22}& \underline{60.64}&\underline{42.80}&\underline{70.64}&\underline{66.97}\\
\hline
\end{tabular}
\caption{Performance comparison on multi-class classification on PDTB 2.0 and PDTB 3.0. Scores with {\bf bold}/\underline{underline} indicate the top/second best performance. {\small $\dagger$} and {\small $\ddagger$} indicate scores are borrowed from Xiang et al. ~\cite{Xiang_etal_2022_Encoding} and Wu et al.~\cite{Wu_etal_2022_A}, respectively. {\small $\bullet$} denotes that we run their source code to obtain the results on PDTB 3.0.}
\label{tab:multi_class}
\end{table*}

\begin{table}
\centering
\small
\begin{tabular}{l|c|c|c|c}
\hline
\bf Model & \bf Temp. & \bf Cont. & \bf Comp. & \bf Exp.\\
\hline
\multicolumn{5}{c}{\bf PDTB 2.0}\\
\hline
Ruan et al.~\cite{Ruan_etal_2020_Interactively}&39.35&59.56 &46.75 &75.83\\
\hline
Liu et al.~\cite{Liu_etal_2020_On} &\underline{50.26}&60.98&	59.44&	77.66 \\
\hline
Dou et al.~\cite{Dou_etal_2021_CVAE}&44.01 &\textbf{63.39}&	55.72&	\textbf{80.34}\\
\hline
Baseline (B) &48.82&60.00&\underline{61.59}& 75.60 \\
B + CMCL& \textbf{53.44}& \underline{62.87}& \textbf{62.34} & \textbf{77.81}\\
\hline
\multicolumn{5}{c}{\bf PDTB 3.0}\\
\hline Ruan et al.~\cite{Ruan_etal_2020_Interactively}{\small $^\dagger$}& 34.74 &61.95& 30.37 &64.28\\
\hline
Liu et al.~\cite{Liu_etal_2020_On}{\small $^\bullet$}&\underline{54.27}&\underline{72.92}&56.95&\underline{73.51} \\
\hline
Xiang et al.~\cite{Xiang_etal_2022_Encoding} & 42.13 &66.77& 35.83 &70.00\\
\hline
Baseline (B) & 52.70& 72.43 &\textbf{59.39}& 72.82 \\
B + CMCL&\textbf{56.17}& \textbf{73.61}& \underline{59.24} & \textbf{75.03}\\
\hline
\end{tabular}
\caption{Performance comparison of multiple binary classification on the top-level classes on PDTB 2.0 and PDTB 3.0.}
\label{tab:binary_class}
\end{table}

\subsection{Experimental Results}

\subsubsection{Multi-class Classification}
Table~\ref{tab:multi_class} presents the comparison results of our approach against representatives of previous studies on multi-class classification. As the results shown, our approach (B + CMCL) achieves competitive performance with substantial improvement over the baseline on both PDTB 2.0 and PDTB 3.0. Specifically, for 4-way classification, our CMCL approach outperforms the baseline by $3.06\%$ in accuracy and $3.03\%$ in F1 score on PDTB 2.0, and by $2.40\%$ in accuracy and $2.71\%$ in F1 score on PDTB 3.0. It should be noted that our baseline model is a simplified version of the model in~\cite{Liu_etal_2020_On}. However, with the proposed CMCL, our approach outperforms~\cite{Liu_etal_2020_On} with absolute better performance on both PDTB 2.0 and PDTB 3.0.

Regarding 11-way classification, unsurprisingly our CMCL approach outperforms the baseline in both accuracy and F1. It also achieves higher performance than Liu et al. ~\cite{Liu_etal_2020_On}. Compared with the latest state-of-the-art model of Wu et al.~\cite{Wu_etal_2022_A} on PDTB 2.0, our CMCL approach achieves $0.31\%$ and $2.31\%$ gains in accuracy and F1, respectively. 

Overall, our approach outperforms all previous work except Long et al.~\cite{Long-etal-2022-facilitating}. The work uses a big batch size (e.g., 256) to ensure that there exists positive pairs especially for instances of infrequent senses. Thus it highly relies on high performance computing to cater such big batch size, which is almost ten times of ours. Besides, this work view instances having same parent discourse relation as positives and instances not having same parent discourse relation as negatives --- this is orthogonal to our main aim of applying CL at different layers and can be incorporated in future work.

\subsubsection{Binary Classification}
Table~\ref{tab:binary_class} presents the results of binary classification on both PDTB 2.0 and PDTB 3.0. As the results shown, compared to the baseline, our CMCL approach achieves better performance for all classes with the only exception of \textit{Comp.} on PDTB 3.0. Moreover,\ignore{ except Long et al.~\cite{Long-etal-2022-facilitating}} our CMCL model achieves the top performance for six binary classification tasks and the second-best performance for the other two tasks. 

Taking PDTB 2.0 as example, on average our CMCL model outperforms Liu et al.~\cite{Liu_etal_2020_On} by $2.03\%$ F1 scores and the baseline by $2.61\%$. It also achieves better performance than the recent  state of the art~\cite{Dou_etal_2021_CVAE} by $3.25\%$ F1 scores.

\section{Analysis}
We take PDTB 2.0 as a representative to discuss how our proposed approach improves performance. 

\subsection{Ablation Study}

\begin{table*}[!t]
\centering
\small
\begin{tabular}{l|l|cc|cc|c|c|c|c||c|c}
\hline
\multirow{2}{*}{\bf \#} & \multirow{2}{*}{\bf Model} & \multicolumn{2}{c|}{\bf 4-way} & \multicolumn{2}{c|}{\bf 11-way}  &\multirow{2}{*}{\bf Temp.}&\multirow{2}{*}{\bf Cont.}&\multirow{2}{*}{\bf Comp.}&\multirow{2}{*}{\bf Exp.}&\bf Para.&\bf Speed\\
\cline{3-6}
& & \bf Acc & \bf F1 & \bf Acc & \bf F1 & & & & &(M)&(min)\\
\hline
1 & Baseline (B) & 67.59 & 62.19 & 56.98 & 39.59 & 48.82 & 60.00 & 61.59 & 75.60 & 1.77 &0.73 \\
2 & B + CMCL (LICL) & \textbf{70.65} & \textbf{65.22} & \textbf{60.64} & \underline{42.80} & \textbf{53.44} & 62.87 & 62.34 & \textbf{77.81} & 3.02&1.12\\
\hline
\hline
3 & B + $\text{LICL}^1$ &69.98& \underline{64.26} &59.77 & \textbf{42.86} &\underline{52.71}& 61.88&\textbf{62.75} & \underline{77.40}&2.96 &0.90 \\
4 & B + $\text{LICL}^2$  &69.12 &62.94& 59.29& 40.85&47.06 & 60.14& 61.86& 76.71& 1.81&0.90 \\  
5 & B + $\text{LICL}^3$&69.69 & 62.00 &	58.33& 38.63 & 49.60 & 61.46&	61.43& 77.18 & 1.81&0.90\\
6 & B + $\text{LICL}^{1\sim 3}$ & 70.17 &64.19&	59.29 &41.64 &48.00& 62.98& \underline{62.58}& 77.31 & 3.02&1.05\\
\hline
\hline
7 & B + $\text{CE}^{1\sim 3}$ &68.36 &61.54 & 58.71 &\underline{42.80}& \textbf{53.44} & \underline{63.52} & 61.88 & 76.06& 1.78& 0.77\\
\hline
\hline
8 & B + CMCL (LCL) &\underline{70.55}&	64.16& 58.33& 40.91 &48.78& \textbf{63.58}& 62.24& 76.48 &3.02&1.10\\
\hline
\end{tabular}
\caption{Ablation study on PDTB 2.0.}
\label{tab:ablation}
\end{table*}

We carry out a series of ablation study and the results are presented in Table~\ref{tab:ablation}. We also compares the parameter sizes and training speed of different models. Compared to the baseline model, our approach introduces additional parameters {\small $W^{\mu}$} in Eq.~\ref{equ:mu} to obtain an instance's representation from its DU1 and DU2. Including CL losses also slightly increases training time, as shown in Table~\ref{tab:ablation}. It is worth noting that our approach only introduce additional parameters and consume additional training time in training, it has same size of parameter and similar decoding time in inference. 

\paragraph{Effect of LICL Applied to Different Layers.} As in Figure~\ref{fig:model_overview}, LICL could be applied on the representation learning of either the context representation layer ({\small $\text{LICL}^1$}), the fusion layer ({\small $\text{LICL}^2$}), or the aggregation layer ({\small $\text{LICL}^3$}). As shown in \#3 $\sim$ \#7 in Table~\ref{tab:ablation}, we have the following observations. 
\begin{itemize}
\item First, applying LICL to any layer achieves better performance than the baseline model.
\item Second, applying LICL to the lowest layer (i.e., {\small $\text{LICL}^1$}) achieves better performance than applying at a higher layer (i.e., {\small $\text{LICL}^2$} or {\small $\text{LICL}^3$}), which illustrates that it is more helpful to constrain the representation learning with LICL in a lower layer than in a higher layer. Note that since we do not fine-tune RoBERTa and word embeddings, applying LICL to the context representation layer (i.e., {\small $\text{LICL}^1$}) will optimize the segment embeddings, which play a critical role \cite{Liu_etal_2020_On}.
\item Third, since there may exist redundancy, simply applying LICL in all the three layers (i.e., {\small $\text{LICL}^{1\sim 3}$}) does not lead to the best performance. Comparing \#3 and \#6, there is no obvious advantage when further applying LICL in the fusion layer and the aggregation layer. 
\item Fourth, from the results of \#2 and \# 6, we observe that our approach achieves higher performance, which suggests that applying CMCL in higher layers is more helpful than directly applying CL in them.
\end{itemize}

\paragraph{CMCL VS. Cross Entropy Losses.} Besides CL losses, it is also practicable to directly apply cross entropy (CE) loss on the output of the context representation layer, the fusion layer, and the aggregation layer. Comparing \#1 and \#7, we observe that adding multiple CE losses improves classification performance over the baseline for all tasks. However, adding CE losses is less effective than adding LICL losses when comparing \#2 and \#7, especially for the 4-way and 11-way classification. 

\paragraph{LCL VS. LICL.} In Section \ref{sect:licl}, we extend LCL to LICL by integrating instance-centered contrastive learning (ICL). To evaluate the contribution of ICL, we make a contrastive study by replacing LICL back with LCL. As shown in \#2 and \#8, LICL achieves better performance than LCL for all classification tasks with the only exception of the binary classification task of \textit{Cont}. This suggests that it is helpful to couple label-centered CL and instance-entered CL into a united one. 

\subsection{Comparison with Different Configurations of Negatives}
\label{sect:experiment_settings}

Rather than defining {\small $\text{Neg}_{\text{ICL}}$} as Eq.~\ref{equ:neg2}, alternatively we can also define it as:
\begin{equation}
\small
\text{Neg}_{\texttt{ICL}}\left(i\right)=\sum_{c\in C\&\&c\neq y_i}\exp\left(Sim\left(L_{c},\mu_{x_i}\right)\right),
\label{equ:neg3}
\end{equation}
which considers all negative labels. Replacing Eq.~\ref{equ:neg2} with Eq.~\ref{equ:neg3} will influence the results of multi-class classification while it does not change the results of binary classification. 

Table~\ref{tab:multi_class_appendix} compares the performance on the PDTB 2.0 test set when {\small $\text{Neg}_{\text{ICL}}$} is defined as either Eq.~\ref{equ:neg2} or Eq.~\ref{equ:neg3}. It shows that for both 4-way and 11-way classification, replacing Eq.~\ref{equ:neg2} with Eq.~\ref{equ:neg3} significantly hurts the performance in both accuracy and Macro F1.

\begin{table}
\small
\centering
\begin{tabular}{l|cc|cc}
\hline
\multirow{2}{*}{\bf {\small $\text{Neg}_{\text{ICL}}$}}& \multicolumn{2}{c|}{\bf 4-way} &\multicolumn{2}{c}{\bf 11-way}\\
\cline{2-5}&  \bf Acc & \bf F1 & \bf Acc & \bf F1 \\
\hline
 Eq.\ref{equ:neg3} & 68.74 & 62.17 & 58.33 & 40.28 \\
\hline
Eq.\ref{equ:neg2} & \textbf{70.65} & \textbf{65.22} & \textbf{60.64} & \textbf{42.80}\\
\hline
\end{tabular}
\caption{Performance comparison on multi-class classification on PDTB 2.0 with different {\small $\text{Neg}_{\text{ICL}}$} definition.}
\label{tab:multi_class_appendix}
\end{table}

\subsection{Effect of Hyper-parameter $\lambda$}
\begin{figure*}[!t]
\centering
\includegraphics[width=6in]{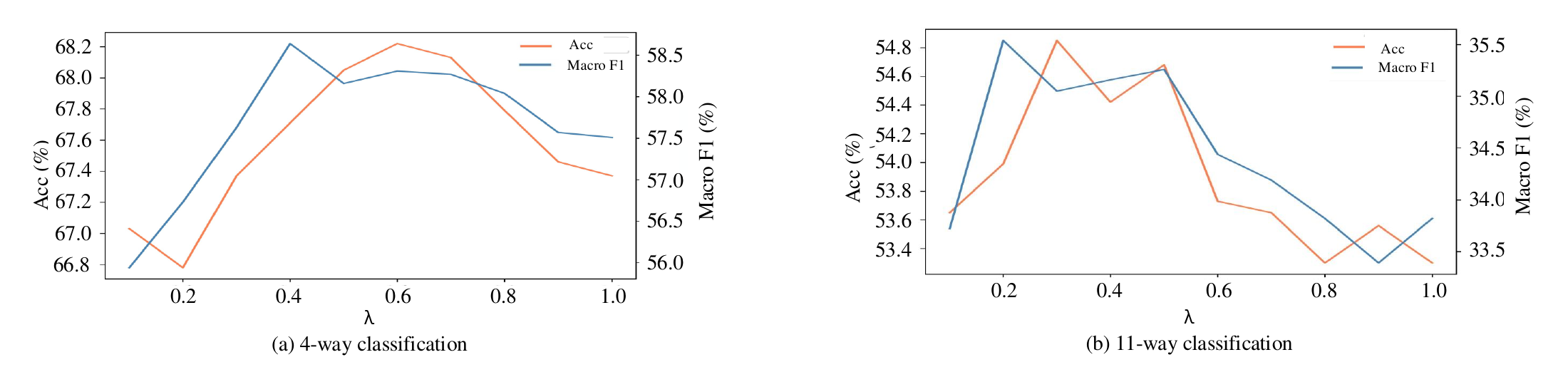}
\caption{Effect of hyper-parameter $\lambda$ on the PDTB 2.0 development set.}
\label{fig:hyperpara_lambda}
\end{figure*}

We use hyper-parameter $\lambda$ to control the contribution of {\small $\text{LICL}^{1}$} loss in the joint training objective in Eq.~\ref{equ:final_loss}. Figure~\ref{fig:hyperpara_lambda} presents the performance tendency of 4-way and 11-way classification tasks on the development set of PDTB 2.0 when $\lambda$ ranges from 0.1 to 1.0 with a step 0.1. Overall, along the increment of $\lambda$, the performance starts to increase at the beginning and then starts to decrease once it reaches the best performance. Specifically, for 4-way classification, it reaches the best performance when $\lambda$ is 0.6 for accuracy and 0.4 for Macro F1. However, for 11-way classification, it gets the best performance when $\lambda$ is 0.3 and 0.2 for accuracy and Macro F1, respectively. In final, we set $\lambda$ as 0.4 and 0.3 for 4-way and 11-way classification on PDTB 2.0, respectively.

\subsection{Representation Visualization of Different Layers}

\begin{figure*}
\centering
\includegraphics[width=5.5in,trim={0.1cm 0.6cm 0.4cm 0.1cm}, clip]{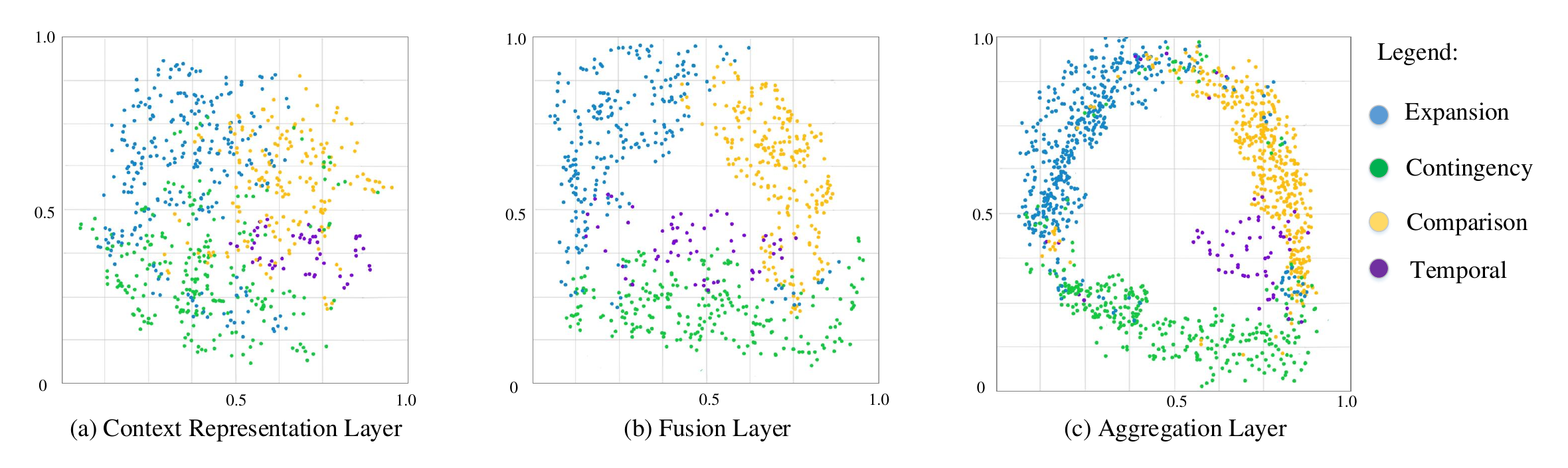}
\caption{Visualization on representations of our approach on PDTB 2.0 development set using T-SNE.}
\label{fig:CMCL_tsne}
\end{figure*}

Figure~\ref{fig:CMCL_tsne} visualizes the representation generated by the three different layers of our CMCL model. As shown in Figure~\ref{fig:CMCL_tsne} (a), 
instances of \textit{Cont.} and \textit{Comp.} tend to be well separated while the instances of \textit{Exp.} and \textit{Temp.} have big overlap with those of \textit{Cont.} and \textit{Comp.}, respectively. In Figure~\ref{fig:CMCL_tsne} (b), we observe that the instances of \textit{Exp.} tend to cluster together, being separated from instances of \textit{Cont}. Finally, in Figure~\ref{fig:CMCL_tsne} (c), instances of \textit{Exp.}, \textit{Cont.}, and \textit{Comp.} are well organized in different corners while instances of \textit{Temp.} locate among the instances of \textit{Cont.} and \textit{Comp}.

Comparing visualization of the baseline model in Figure~\ref{fig:baseline_tsne}, we can conclude that the representations from the proposed CMCL model are more discriminative than those from the baseline model, indicating that our approach is effective to enhance representation learning and improve capability of discriminating instances of different classes. 
 
\section{Related Work}

\subsection{Implicit Discourse Relationship Recognition}
Existing approaches to IDRR mainly put effort on three directions. The first direction is to enhance the representation of DUs. For example, knowledge-enhanced representation learning~\cite{Guo_etal_2020_Working} and entity-enhanced representation learning~\cite{Shi_etal_2021_Entity} are explored in this direction. Bai et al.~\cite{Bai_etal_2018_Deep} use ELMo to build multi-grained embedding layers and improve 4-way classification to above $50\%$ F1 score. \cite{He_etal_2020_Transs} model discourse relations by interpreting them as translations operating in the low-dimensional embedding space and use the learned semantic information to enhance the representation of DUs. Recent studies~\cite{Jiang-etal-2022-global,Long-etal-2022-facilitating} have also applied contrastive learning for IDRR. Different from ours, they take advantage of the three-level sense hierarchy in PDTB ranging from four broad top-level senses to more specific senses below them. Specifically, they view instances having same parent discourse relation as positives and instances not having same parent discourse relation as negatives. \ignore{Besides, Jiang et al.~\cite{Jiang-etal-2022-global} obtain label embeddings through a graph neural network while we initialize label embeddings randomly. Long et al.~\cite{Long-etal-2022-facilitating} use a big batch size (e.g., of 256 instances) to ensure that there exists at least one positive in the same batch, especially for an instance of infrequent sense. \textcolor{red}{Note that although Long et al.~\cite{Long-etal-2022-facilitating} achieves the new state of the art, it highly relies on high performance computing to cater the big batch size, which is almost ten times of ours. }}The second direction aims to capture semantic interaction between DUs. Various modules with attention mechanism are proposed in this direction as attention-based modules have strong capability of learning interaction information between DUs~\cite{Chen_etal_2016_Implicit,Lan_etal_2017_Multi,Guo_etal_2018_Implicit,Liu_etal_2020_On,Ruan_etal_2020_Interactively,Xiang_etal_2022_Encoding}. The third direction is to jointly learn IDRR with other relevant tasks under multi-task learning framework. One example of multi-task learning is to combine a connective classifier together with the regular relation classifier~\cite{Bai_etal_2018_Deep,Nguyen_etal_2019_Employing, Wu_etal_2020_Hierarchical,Wu_etal_2022_A}. The idea is inspired by the fact that connective classification and discourse relation classification are two closely relevant tasks and could help to improve each other under the multi-task learning framework. To elicit knowledge from large pre-trained models, recent studies \cite{Xiang-etal-2022-connprompt,Zhou-etal-2022-prompt-based} also propose prompt learning for IDDR, which is viewed as a connective prediction task. However, they need to fine-tune the pre-trained models, thus significantly increase the size of model parameter. Our CMCL approach also belongs to the scope of representation enhancement. Unlike all previous IDRR studies in this direction, we propose to adapt label- and instance-centered CL to enhance representation learning, and further propose a novel constrained multi-layer CL approach to properly impose a constraint that the contrastive loss of higher layers should be smaller than that of lower layers. 

\subsection{Label-aware Contrastive Learning}
Contrastive learning can be divided into two categories: self-supervised CL and  supervised CL. Self-supervised CL contrasts a single positive for each anchor against a set of negative instances from the entire remainder of the batch. Supervised CL contrasts the set of instances from the same class as positives against the negative instances from the remainder of the batch. Another important difference between self-supervised CL and supervised CL is that supervised CL can construct positives by leveraging label information~\cite{Gunel_etal_2021_Supervised,Suresh_etal_2021_Not,Zhang_etal_2022_Label}. For example, Suresh et al.~\cite{Suresh_etal_2021_Not} propose label-aware 
supervised CL which assigns variously weights to instance pairs based on labels. Zhang et al.~\cite{Zhang_etal_2022_Label} propose a label-anchored supervised CL approach which leverages label embeddings to take the place of data augmentation. 

\section{Conclusion}
In this paper we build an IDRR system with relatively simple neural network. For better representation learning for DUs, we adapt label-centered contrastive learning proposed by \cite{Zhang_etal_2022_Label} and propose label- and instance-centered contrastive learning (LICL), which can be plugged into each inside layers of the IDRR model. Finally, we propose a novel multi-layer contrastive learning approach, named constrained multi-layer contrastive learning (CMCL) to properly constrain that the LICL loss of higher layers should be smaller than that of lower layers. Extensive experimental results on two benchmarks show that our approach achieves significant improvement over the baseline. This is very encouraging since in inference phase the approach does not introduce additional parameters and sophisticated neural network modules.

\bibliography{ecai}
\end{document}